\documentclass[11pt,a4paper]{article}
\usepackage[hyperref]{acl2021}
\usepackage{times}
\usepackage{latexsym}
\usepackage{graphicx}
\usepackage{caption}
\usepackage{subcaption}
\usepackage{float}
\usepackage{booktabs}
\usepackage{amsmath}
\usepackage{bm}
\usepackage{multirow}
\usepackage{amssymb}
\usepackage{stmaryrd}
\usepackage{tabulary}

\usepackage[T1]{fontenc}

\usepackage[utf8]{inputenc}

\usepackage{microtype}

\newcommand{\eg}{\textit{e.g.}}
\newcommand{\ie}{\textit{i.e.}}
\newcommand{\alg}{attention modulation}

\aclfinalcopy 

\title{
On-the-Fly Attention Modulation for Neural Generation
}

\author{Yue Dong$^1$\thanks{*This work was done when the first author was
an intern at AI2.} \quad Chandra Bhagavatula$^2$ \quad Ximing Lu$^{2,4}$ \quad Jena D. Hwang$^2$ \\\quad \textbf{Antoine Bosselut}$^{2,3}$ \quad
 \textbf{Jackie Chi Kit Cheung}$^1$  \quad \textbf{Yejin Choi}$^{2,4}$\\ \\
    $^1$Mila / McGill University \quad $^2$Allen Institute for Artificial Intelligence\\
    $^3$ Stanford University \quad $^4$Paul G. Allen School of CSE, University of Washington\\
    {\small \{\tt yue.dong2@mail, jcheung@cs\}.mcgill.ca} \\
    \small \{\tt  chandrab,ximinglu,jenah,antoineb,yejinc\}@allenai.org    }
    
\begin{document}
 \maketitle
\begin{abstract}
Despite considerable advancements with deep neural language models (LMs), neural text generation still suffers from \textit{de}generation: the generated text is repetitive, generic, self-contradictory, and often lacks commonsense. Our analyses on sentence-level attention patterns in LMs reveal that neural degeneration may be associated with insufficient learning of task-specific characteristics by the attention mechanism.  This finding motivates on-the-fly \alg{}\protect\footnotemark-- a simple but effective method that enables the injection of priors into attention computation during inference. Automatic and human evaluation results on three text generation benchmarks demonstrate that \alg{} helps LMs generate text with enhanced fluency, creativity, and commonsense reasoning, in addition to significantly reduce sentence-level repetition. 
\end{abstract}

\section{Introduction}

Neural text generation is critical for a wide range of downstream natural language applications. However, the standard approach -- using a Transformer-based \citep{Vaswani2017AttentionIA} language model (\eg, \citealp{radford2019language}) with maximum likelihood fine-tuning and non-stochastic decoding  – is known to exhibit \textit{de}generation \citep{welleck2020neural}. Despite being pre-trained on large amounts of data, text generated by neural models is observed to be repetitive, generic, self-contradictory, and lacking commonsense \citep{holtzman2020curious}.

Many explanations have been proposed for neural text degeneration, including inappropriate training objectives \citep{welleck2020neural} and decoding discrepancies relative to human language \cite{holtzman-etal-2018-learning, holtzman2020curious}.  While the aforementioned may be factors for neural degeneration, we show that insufficient learning of \textit{task-specific} characteristics -- reflected in the self-attention mechanism in transformer blocks -- is associated with neural text degeneration. We demonstrate that degeneration is alleviated if we inject priors through \alg{} (AttnM) 
during inference.
\begin{figure}[t]
    \centering
    \includegraphics[scale=0.52]{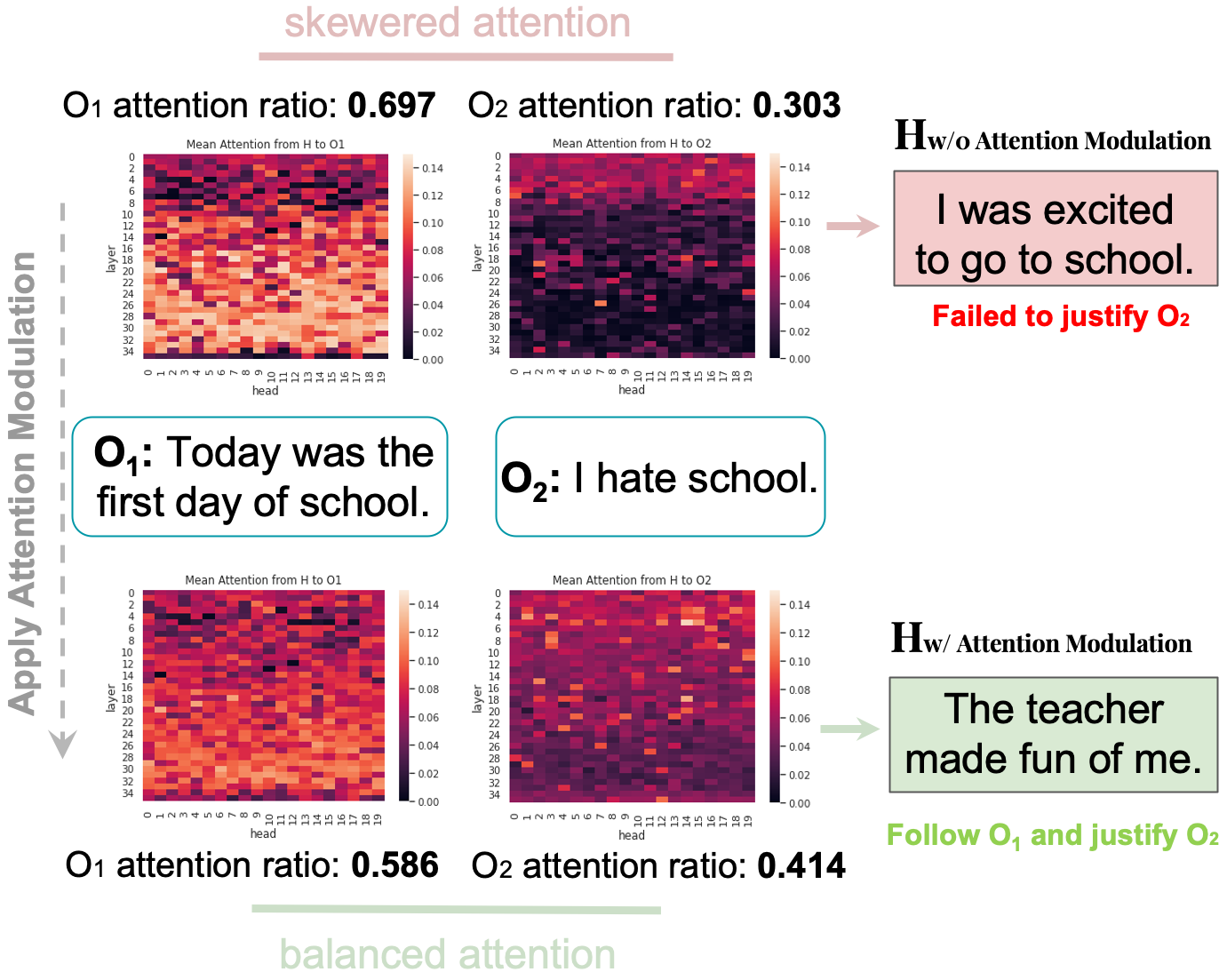}

    \caption{Example of fine-tuned GPT2-L outputs without (top) and with (bottom)  \alg{} on $\alpha$NLG. The task is to generate a plausible explanatory hypothesis $H$ for observations $O_1$ and $O_2$.  Our proposed \alg{} injects the task-specific prior -- LMs should consider both observations relative equally --  through balancing the sentence-level attention weights (Eqn. \ref{eq:mean_sent_to_sent_attention}) in Transformer blocks during inference. Applying \alg{} with the aforementioned prior make sentence-level attentions from generation to observation pairs ($O_1$,$O_2$) more balanced\protect\footnotemark, which are reflected in the sentence-level attention heatmaps of GPT2-L (darker = lower attention) across layers (y-axis) and heads (x-axis). 
    }
\end{figure}

\footnotetext{attention ratios are normalized mean sentence-to-sentence attention from generation $H$ to observations $O_1$ and $O_2$}

Self-attention -- the  ubiquitous component of Transformers -- is task-agnostic with a large learning capacity for many NLP tasks \citep{Vaswani2017AttentionIA,devlin-etal-2019-bert,Brown2020LanguageMA}. It learns the general characteristics of language processing through pre-training on large amounts of unlabeled data. For example, multiple analyses have suggested that attention patterns in pre-trained Transformers implicitly encode syntactic information \citep{raganato-tiedemann-2018-analysis,NEURIPS2019_2c601ad9, vig-belinkov-2019-analyzing}. 
In sequence transduction tasks, these learned characteristics, embedded in attention,  make pre-trained Transformers a powerful language model \citep{radford2019language}.  

A final task-specific step is typically required for adapting a task-agnostic language model to perform the desired task\footnote{\citet{Brown2020LanguageMA} have shown that GPT3 greatly improves task-agnostic, few-shot performance, but still struggles on tasks with strong task-specific characteristics.}. However, these task-specific characteristics might not sufficiently coincide with general characteristics even after fine-tuning. 
For example, task-specific characteristics embedded in attention patterns -- such as word alignments for machine translation -- are often noisy and imperfect for generalization \citep{kobayashi-etal-2020-attention}. 

We show that insufficient learning of task-specific characteristics, reflected in sentence-level 
attention patterns\footnote{We study the global context in the multi-sentence prompts and choose sentence-level attention (Eqn. \ref{eq:agg_sent_to_sent_attention}) as the experiment unit, since sentences are linguistic units of complete meaning.}  often being out of focus, may be associated with neural text degeneration (\S \ref{sec:motivation}). Based on this observation, we propose a simple attention modulation framework that can dynamically redistribute sentence-level attention weights by injecting task-specific priors in Transformer blocks for different downstream tasks (\S \ref{sec:model}). Remarkably, in long-range narrative story generation, abductive reasoning generation and constrained commonsense text generation, both automatic and human evaluation have shown improved quality in fluency, dullness, repetition, and commonsense reasoning with \alg{} (\S \ref{sec:result}).

\section{Background}\label{sec:vinalla_attention}
We briefly discuss how vanilla attention works, as well as Transformer architecture used in this paper. 

\paragraph{Single-headed attention}
Given a sequence of $d$-dimensional input vectors  $\bm{x} =\{ \bm{x}_1,\ldots,\bm{x}_n \}$, attention mechanism computes a set of weights based on a query vector $\bm{y}_i \in \mathbb{R}^d$:
\begin{equation}
\label{eq:single_attention_vector}
  \text{Attn}(\bm{x},\bm{y}_i) =  (\alpha_{i,1}(\bm{x},\bm{y}_i),\ldots,\alpha_{i,n}(\bm{x},\bm{y}_i))
\end{equation}
where $\alpha_{i,j}$ is the attention weight that $\bm{y}_i$ pays to $\bm{x}_j$. One formulation of attention --- scaled dot product attention --- is computed as:
\begin{equation}
\label{eq:single_attn_weight}
    \alpha_{i,j} := \underset{\bm{x}_j \in \bm{x}}{\text{softmax}}
    \Big(\frac{\bm{q}(\bm{y}_i) \bm{k}(\bm{x}_j)^\top}{\sqrt{d}} \Big) \in \mathbb{R}
\end{equation}
where query  $\bm{q}(\cdot)$ and key  $\bm{k}(\cdot)$ functions are linear transformations.  In self attention, every $\bm{x}_i$ is used as the query vector ($\bm{y}_i$). An updated representation $\tilde{\bm{x}}_i$ is computed as a weighted sum of value vectors that are linearly transformed by $\bm{v}(\cdot)$:
\begin{equation}
\label{eq:single_attention_update}
    \tilde{\bm{x}}_i= \sum_{\bm{x}_j \in \bm{x}} \alpha_{i,j}\bm{v}(\bm{x}_j).
\end{equation} 

\paragraph{Multi-head attention} In \textit{multi-headed} attention (MHA), $N_h$   attention heads are computed independently to obtain the updated $\tilde{\bm{x}}_i$: 
\begin{equation}
\label{eq:multi_head_attention}
    \tilde{\bm{x}}_i = W_o \bigparallel_{h=1}^{N_h} \Big( \sum_{\bm{x}_j \in \bm{x}} \alpha^h_{i,j}\bm{v^h}(\bm{x}_j)\Big).
\end{equation}
$\alpha^h_{i,j}$ follows Eqn. \ref{eq:single_attn_weight} except the model dimension in each head $h$ is often reduced to $d_h=\frac{d}{N_h}$. 
 $\tilde{\bm{x}}_i$ is obtained by the concatenation of lower-rank representations from all heads and $W_o \in \mathbb{R}^{  d  \times h \cdot d_h}$. 

\paragraph{GPT2-L} GPT2 \citep{radford2019language} is a family of Transformer-based language models (LMs) that follows the architecture of stacked decoder.  
As GPT2 follows a multi-layer and multi-headed setting, $\alpha_{i,j}$ is specific to a layer $l$ and head $h$, noted as $\alpha_{i,j}^{l,h}$. We use the GPT2-L model that has 36 layers with 20 heads per layer (762M total parameters).

\section{Neural text degeneration vs. attention}\label{sec:motivation}
As researchers have sought to understand the internal mechanisms of Transformers, the attention patterns exhibited by these heads have drawn considerable study \citep{vig-belinkov-2019-analyzing,jain2019attention,wiegreffe2019attention}. We perform sentence-level attention analysis to explore whether aggregated attention patterns are associated with neural text degeneration. 

\subsection{Sentence-level attention}  We first define the sentence-to-sentence attention of a language model $\mathcal{M}$ with $L$ layers and $H$ heads. Given two sentences $p$ and $g$ such that $p$ precedes $g$, the mean $\bar{\alpha}^{l,h}_{g,p}$ and max $\hat{\alpha}^{l,h}_{g,p}$ sentence-to-sentence attentions from  $g$ to  $p$ for layer $l$ and head $h$ are: 

\begin{equation} 
\label{eq:mean_sent_to_sent_attention}
      \bar{\alpha}^{l,h}_{g,p} =\frac{\sum\limits_{i=1}^{|g|}\sum\limits_{j=1}^{|p|}\alpha^{l,h}_{i,j}(g_i,p_j)}
    {|g| \cdot |p|}
\end{equation}

\begin{equation}
\label{eq:max_sent_to_sent_attention}
     \hat{\alpha}^{l,h}_{g,p} =\max_{\substack{i \in \{1,\ldots,|g|\} \\ j \in \{1,\ldots,|p|\}}}\alpha^{l,h}_{i,j}(g_i,p_j).
\end{equation}

The aggregated sentence-to-sentence attention over the Transformer architecture $\mathcal{M}$ is defined as:
\begin{equation}
\label{eq:agg_sent_to_sent_attention}
      \alpha^{\mathcal{M}}_{g,p} = \frac{\sum\limits_{l=1}^{L}\sum\limits_{h=1}^{H}\alpha^{l,h}_{g,p}}
    {L \cdot H}
\end{equation}
where $\alpha \in \{\bar{\alpha},\hat{\alpha}\}$ computes either the mean or the max sentence-level attention over $\mathcal{M}$. 

\subsection{Is neural text degeneration related to attention patterns?}\label{subsec:motivation_degeneration}
We conduct experiments to evaluate whether neural text degeneration is associated with sentence-level attention patterns. Empirical results on two types of neural degeneration that are easy to detect -- repetition and lacking commonsense reasoning under constraints -- reveal their association.

\paragraph{Repetition vs. attention}
One common form of neural text degeneration is sentence-level repetition \citep{welleck2020neural}. This type of degeneration happens frequently in our experiment on ROCStories test set (\S \ref{subsec:result_roc}):
given a five-sentence prompt, 35.4\% of the consecutive sentences from the next five greedily generated sentences by the fine-tuned GPT2-L are exact repetitions. We check whether sentence-level attention patterns behave differently when generating repeated or different consecutive sentences.  

\begin{figure}[t]
\centering
\includegraphics[scale=0.55]{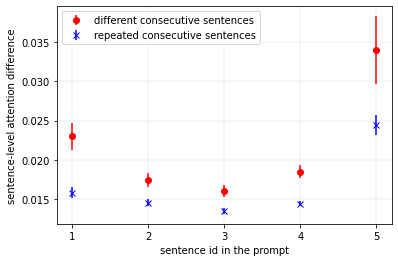}
\caption{Mean sentence-level attention change of GPT2-L on ROCStories test set, while generating different (red) or repeated (blue) consecutive sentences.}
\label{fig:sentence_level_attention_repetition}
\end{figure} 

We inspect the attention behavior by measuring the change of sentence-level attention when generating two consecutive sentences.  
The generations of fine-tuned GPT2-L on ROCStories test set are separated into two subsets $\{\mathcal{D}_{\text{repeated}},\mathcal{D}_{\text{different}} \}$: in which consecutive sentences that are either repeated (\ie, degenerate) or different. Given the fine-tuned GPT2-L language model $\mathcal{M}$, we measure the sentence-level attention change $\Delta$ on the prompt sentence $p_{j \in \{ 1, \ldots, 5\}}$ while generating the consecutive sentence pair $(g_i, g_{i+1}),i \in \{1,\ldots, 4\}$, aggregated over the subset $\mathcal{D} \in \{\mathcal{D}_{\text{different}},  \mathcal{D}_{\text{repeated}}\}$: 

\begin{equation}
\label{eq:sentence_attention_change}
    \Delta (j,\mathcal{D}) = \sum_{d \in \mathcal{D}}|\bar{\alpha}^{\mathcal{M}}_{g^d_{i+1},p^d_{j}} - \bar{\alpha}^{\mathcal{M}}_{g^d_{i},p^d_{j}}|/|\mathcal{D}|
\end{equation}
where $\bar{\alpha}^{\mathcal{M}}_{g_{i},p_{j}}$ is the mean sentence-level attention from sentence $g_{i}$ to sentence $p_{j}$ defined in Eqn. \ref{eq:agg_sent_to_sent_attention}.

Figure \ref{fig:sentence_level_attention_repetition} plots the aggregated mean sentence-level attention change over prompt sentences when GPT2-L generates repeated (red) or different (blue) consecutive sentences. The sentence-level attention changes are vastly lower on all prompt sentences when generating repeated consecutive sentences. 
Thus, sentence-level repetition may be correlated with the insufficient change of sentence-level attention. In \S \ref{sec:model} and \S \ref{sec:result}, we show that generation quality can be vastly improved by injecting the prior -- attention should look at the prompt differently when generating different sentences -- through our proposed \alg{}.

\paragraph{Lack of commonsense reasoning vs. attention}

Text generated by neural language models is also observed to be lacking commonsense reasoning \cite{mao2019improving}. We check whether this type of neural degeneration is associated with attention patterns. A benchmark dataset for generative commonsense reasoning -- CommonGen \citep{lin-etal-2020-commongen} -- is used as our test bed. CommonGen is designed for constrained commonsense reasoning: given a set of common concepts (e.g., {use, tool, piece, metal}); the task is to generate a coherent and plausible sentence covering all these concepts (e.g., "a piece of metal is used for making tools"). Covering the concepts in generation requires relational reasoning with background commonsense knowledge.

Each concept is represented as a prompt sentence in our experiments.\footnote{We can obtain a clear boundary for each concept with this design choice of adding a period as the separator, as concepts can be tokenized into multiple subwords by GPT2 tokenizers.} During generation, a concept (\eg swim) is covered if its reflected form (\eg \{swim, swimming, swam, swum\}) is generated in the CommonGen test set.  We use a fine-tuned GPT2-L for the generation. Among the 5988 concepts in the prompt, about 75\% of them are covered in the generation of GPT2-L. We can then easily separate sentence-level attention from the generation to the concept into two subsets: concept in the prompt that is covered or uncovered by the generated sentence. 

Table \ref{tab:commongen_analysis} shows the results of max sentence-level attention (Eqn. \ref{eq:agg_sent_to_sent_attention}) of the finetuned GPT2-L on the CommonGen test set\footnote{We measure the max sentence-level attention rather than the mean sentence-level attention on CommonGen, as the attention to a concept is usually reduced once it is generated.}. We can observe that sentence-level attention from the generation to the concept is vastly higher when the concept is covered. Compared to that of uncovered concepts, the aggregated max sentence-level attention is 15.4\% higher. Therefore, failing to generate a common concept through reasoning may be associated with insufficient attention to the concept.   
\begin{table}[t]
\begin{tabular}{l|ccc}
\toprule
            & agg. max attn. & SD     & \#   \\ \midrule
covered     & 0.434                    & 0.0040 & 4515 \\
uncovered & 0.376                    & 0.0068 & 1473 \\ \bottomrule
\end{tabular} 
\caption{Aggregated max sentence-level attention of the fine-tuned GPT2-L; the results are aggregated from the generation to covered or uncovered concepts on the CommonGen test set. agg. max attn., SD, \# refer to aggregated max sentence-level attention, standard deviation, and the number of instances. }\label{tab:commongen_analysis}
\end{table}

In both cases, neural text degeneration is associated with insufficient attention to elements that are important for downstream generations. This motivates us to explore whether we can inject these priors in the language model by altering the attention mechanism to alleviate degeneration.

\section{Method}\label{sec:model}
This section describes our method -- \alg{} -- that can alleviate neural text degeneration. In \S \ref{subsec:model_reweight_attention}, we describe the general \alg{} framework. In \S \ref{subsec:mdoel_roc_reweight}, \S \ref{subsec:mdoel_anlg_reweight}, and \S \ref{subsec:mdoel_commongen_reweight}, we discuss the priors injected through \alg{} for three different tasks: narrative story generation, abductive reasoning generation, and constrained commonsense reasoning. 

\subsection{Attention Modulation} \label{subsec:model_reweight_attention}
Attention modulation aims to change the attention weights of a Transformer-based language model during inference, so that the generation can reflect priors that alleviate neural text degeneration. This additional signal is added to the self-attention computation in the Transformer blocks. 

We reformulate the attention computation of Eqn. \ref{eq:single_attn_weight} by adding an attention reweighting function $f$, where priors can be injected. Given a sequence of input tokens $\bm{x}$, the self-attention from $\bm{x}_i$ to $\bm{x}_j$ ($i \geq j$) while generating the $t$-token is reformulated to:

\begin{equation}
\label{eq:attention_reweight}
    \alpha^t_{i,j} := \underset{\bm{x}_j \in \bm{x}}{\text{softmax}}
    \Big(\frac{\bm{q}(\bm{x}_i) \bm{k}(\bm{x}_j)^\top}{\sqrt{d}} + f_{i,j}(\bm{x}, \bm{\alpha}^{t-1}) \Big) 
\end{equation}
where $f(\bm{x}, \bm{\alpha}^{t-1})$ is the attention reweighting function and $\bm{\alpha}^{t-1}$ is the attention weight matrix for all layers and heads in the Transformer architecture at time step $t-1$. 
The attention reweighting function $f$ can be either pre-defined or learned. In our experiments, we inject pre-defined sentence-level priors (heuristics) through $f$ and show that this injection alleviates neural text degeneration. We leave the learning of better reweighting functions automatically to future work. 

In the following sections, we describe sentence-level attention reweighting functions that are used for three different text generation tasks. 

\subsection{ROCStories: narrative generation}\label{subsec:mdoel_roc_reweight}
As shown in \S \ref{sec:motivation}, sentence repetition in long-form generation may be associated with insufficient attention change while generating consecutive sentences. To amplify the attention changes, we can redistribute sentence-level attention with some priors while generating consecutive sentences. 

We choose the prior that language model should consider long-range context during generation, as we observed that attention mostly focuses on the near history in many cases (Appendix A.1). 
Note this prior also increases the sentence-level attention change while generating consecutive sentences: the sentence-level attention for all previous sentences is always re-balanced based on the newly-generated sentence.
To balance the attention of tokens in each sentence received while generating the next sentence, we define the attention reweight function in Eqn. \ref{eq:attention_reweight} with the aforementioned prior for ROCStories as: 

\begin{equation}
\label{eq:roc_attention_weight}
f_{i,j}(\bm{x}, \bm{\alpha}^{t-1}_{i,j}) = \frac{1}{\alpha^{t-1}_{g_i,p_j}}.
\end{equation}

As later sentences in the prompt usually receive larger sentence-level attention weights (Appendix A.1),  
 attention reweighting function defined in Eqn. \ref{eq:roc_attention_weight} will add a large weight to tokens in the early sentences and a small weight to tokens in the late context sentences. The simple heuristic of balancing context sentences to be considered relatively equal, namely more weights on early context sentences, might not be optimal prior. However, it improves the long-form story generation in multiple measures, including fluency, interesting, newness, and repetition (\S \ref{subsec:result_roc}). 

\subsection{$\alpha$NLG: abductive reasoning generation}\label{subsec:mdoel_anlg_reweight}
The second benchmark dataset we tested with \alg{} is $\alpha$NLG \citep{bhagavatula2020abductive}. This dataset is proposed for abductive reasoning generation: given two observations $O_1$ and $O_2$, the model needs to generate a valid hypothesis $h$ that explains what happened between the two observations. For example, given $O_1$: "Today was the first day of school." and $O_2$: "I hate school.", the task is to generate $h$ such as "The teacher made fun of me." as a plausible explanation.

\citet{bhagavatula2020abductive} has shown that the fine-tuned GPT2 performs far below human performance on the $\alpha$NLG task. We hypothesis that this may be associated with insufficient learning of sentence-level attention to both observations; for example, the model might over-fit to one of the observations for generation. Thus, we inject the prior -- the language model should consider both observations relative equally -- while generating a plausible explanation. This prior can be injected with attention reweighting function defined in Eqn.  \ref{eq:roc_attention_weight}. 

\subsection{CommonGen: constrained commonsense generation}\label{subsec:mdoel_commongen_reweight}
The third benchmark is CommonGen -- a constrained text generation challenge for generative commonsense reasoning. CommonGen requires machines to generate a realistic sentence using \textit{all} concepts from a given concept set by conducting commonsense reasoning over the relations among the given concepts. To successfully generate a plausible and grammatical sentence that follows the commonsense, models need to conduct commonsense reasoning over the relations among the given concepts. 
Our experiment in \S \ref{subsec:motivation_degeneration} shows that the fine-tuned GPT2-L can only cover about $75\%$ of concepts during generation. We infer from Table \ref{tab:commongen_analysis} that this may be associated with GPT2-L giving insufficient sentence-level attention to uncovered concepts. Thus, we propose a simple heuristic -- model should pay more attention to concepts that are not covered yet -- to be injected with \alg{}.

Consider the prompt with $m$ concepts $\bm{c}=\{c_1,\ldots,c_m\}$ and a partially generated sentence $y_1,\ldots,y_{t-1}$. While generating the $t$-th token, the sentence-level reweighting function from the $i$-th token to the $j$-th token in $c_k$ is defined as:
\begin{equation} 
\label{eq:single_attention_weight}
    f_{i,j}(\bm{x}) = 
     \begin{cases}
1/m &\text{if} \quad c_k \subset y_{1:t-1}\\
1 &\text{else}\\
\end{cases}
\end{equation}
Intuitively, if a concept $c_k$ is covered in the partial generation, \alg{} with Eqn. \ref{eq:single_attention_weight} will reduce the attention weights of the tokens in concept $c_k$.

\section{Experimental Setups}\label{sec:experiment}
This section describes the experiment setups, including the baselines, decoding algorithms, datasets, and evaluation metrics.

\paragraph{Model architecture \& baseline } Attention modulation is architecture-agnostic and can be applied to any Transformer-based models that contain self-attention computation. We choose GPT2-L \citep{radford2019language} for our experiments, which has achieved state-of-the-art performance on a variety of generation tasks \citep{vig-belinkov-2019-analyzing}. Attention modulation can be applied to any range of layers in the Transformer. 
To compare models with and without \alg{} on each of the three generation tasks, we use the best fine-tuned GPT2-L based on the validation set after  fine-tuning for 4 epochs with the default settings.

\begin{table*}[t]
\centering
\resizebox{\linewidth}{!}{%
\begin{tabular}{l|rr|rr|rr|rr|rrr}
\toprule
& \multicolumn{2}{c|}{\textbf{next 1 sent.}} & \multicolumn{2}{c|}{\textbf{next 2 sent.}} & \multicolumn{2}{c|}{\textbf{next 3 sent.}} & \multicolumn{2}{c|}{\textbf{next 4 sent.}} & \multicolumn{3}{c}{\textbf{next 5 sent.}}  \\
 & \textbf{uniq.} & \textbf{\% rel.}& \textbf{uniq.} & \textbf{\% rel.} & \textbf{uniq.} & \textbf{\% rel.}  & \textbf{uniq.} & \textbf{\% rel.} & \textbf{uniq.$\uparrow$}  & \textbf{\% rel.$\uparrow$} &\textbf{ \% rep.$\downarrow$} \\ \midrule
w/o AttnM                          & 9.08k     & 48.59   & 13.58k   & 48.78  & 15.71k   & 49.20 & 16.01k   & 49.97  & 17.03k   & 50.52 & 35.43 \\
w/ AttnM  (ours)                            & \textbf{10.39k}   & \textbf{52.92}   & \textbf{15.12k}   & \textbf{54.23}  & \textbf{18.33k}   & \textbf{55.35} & \textbf{20.41k}   & \textbf{55.78}  & \textbf{22.69k}   & \textbf{55.78} & \textbf{17.49}\\
ratio &  1.14     &         & 1.12    &        & 1.16    &       & 1.24    &        & 1.34    &        \\ \bottomrule
\end{tabular}} 
\caption{Test results of the fine-tuned GPT2-L w/ and w/o \alg{} on ROCstories with the greedy decoding algorithm. uniq. represent the unique number of tokens generated in the whole test corpus, which measures the number of new unique tokens generated. rel. represent relevancy, which measures the percentage of tokens generated appears in the prompt. rep. measures the sentence-level repetition -- whether two sentences generated are identical.}\label{tab:roc_result}
\end{table*}

\paragraph{Decoding} Attention modulation directly changes the attention weights of the context tokens during inference. It is orthogonal to different decoding algorithms that change the searching strategies based on the softmax distribution emitted by Transformers.\footnote{These search-based decoding algorithms do not resolve the poorly generated token-level probabilities.} We present the results with non-stochastic decoding algorithms (\ie{}  greedy decoding and beam search), as generations based on them truly reflect the token-level probabilities predicted by the model \citep{holtzman-etal-2018-learning}.

\paragraph{Datasets} We use three different generation datasets -- ROCStories \citep{mostafazadeh-etal-2016-corpus}, $\alpha$NLG \citep{bhagavatula2020abductive}, and CommonGen \citep{lin-etal-2020-commongen}. 
For ROCStories, we used the 2017 version and split the data into 75/10/15 for train/val/test. 

\paragraph{Evaluation} On ROCStories, we measure dullness, relevancy and repetition similar to \citet{welleck2020neural}. We report the number of unique tokens generated, where the generation is less dull if more unique tokens are generated. For repetition, we directly measure sentence-level repetition: two generated sentences are repeated if their strings are the same. For relevancy, we measure the percentage of generated tokens that appear in the prompt. Besides, we perform a human evaluation, where three annotators are asked to rate the generations based on fluency, interestingness, newness, relevancy, and repetition.   

On $\alpha$NLG, we score the generated explanation with respect to the reference using the following automatic metrics: BLEU \citep{Papineni2002BleuAM}, ROUGE \citep{Lin2004ROUGEAP}, METEOR \citep{Banerjee2005METEORAA}, and CIDEr \citep{Vedantam2015CIDErCI}. In addition, we ask annotators to compare the generated explanations without and with \alg{}. Human judges are asked to decide which system provides a more plausible explanation of the observations.

On CommonGen, we report SPICE \citep{anderson2016spice} -- a measure that evaluates semantic propositional content, in addtion to BLEU, ROUGE, METEOR, CIDEr. We also report Coverage \citep{lin-etal-2020-commongen}, which computes the average percentage of input concepts that appear in the lemmatized outputs. We conduct a human evaluation following the protocol of \citet{lu2020neurologic}. Human judges are asked to compare two systems in terms of fluency, coverage (covers the concept), and overall quality (covers the concepts and follows commonsense).

\section{Result}\label{sec:result}
In this section, we present the vast improvements of the fine-tuned GPT2-L with \alg{} on three narrative generation and generative reasoning tasks: ROCStories, $\alpha$NLG, and CommonGen. 

\begin{figure}[t]
\centering
\includegraphics[scale=0.6]{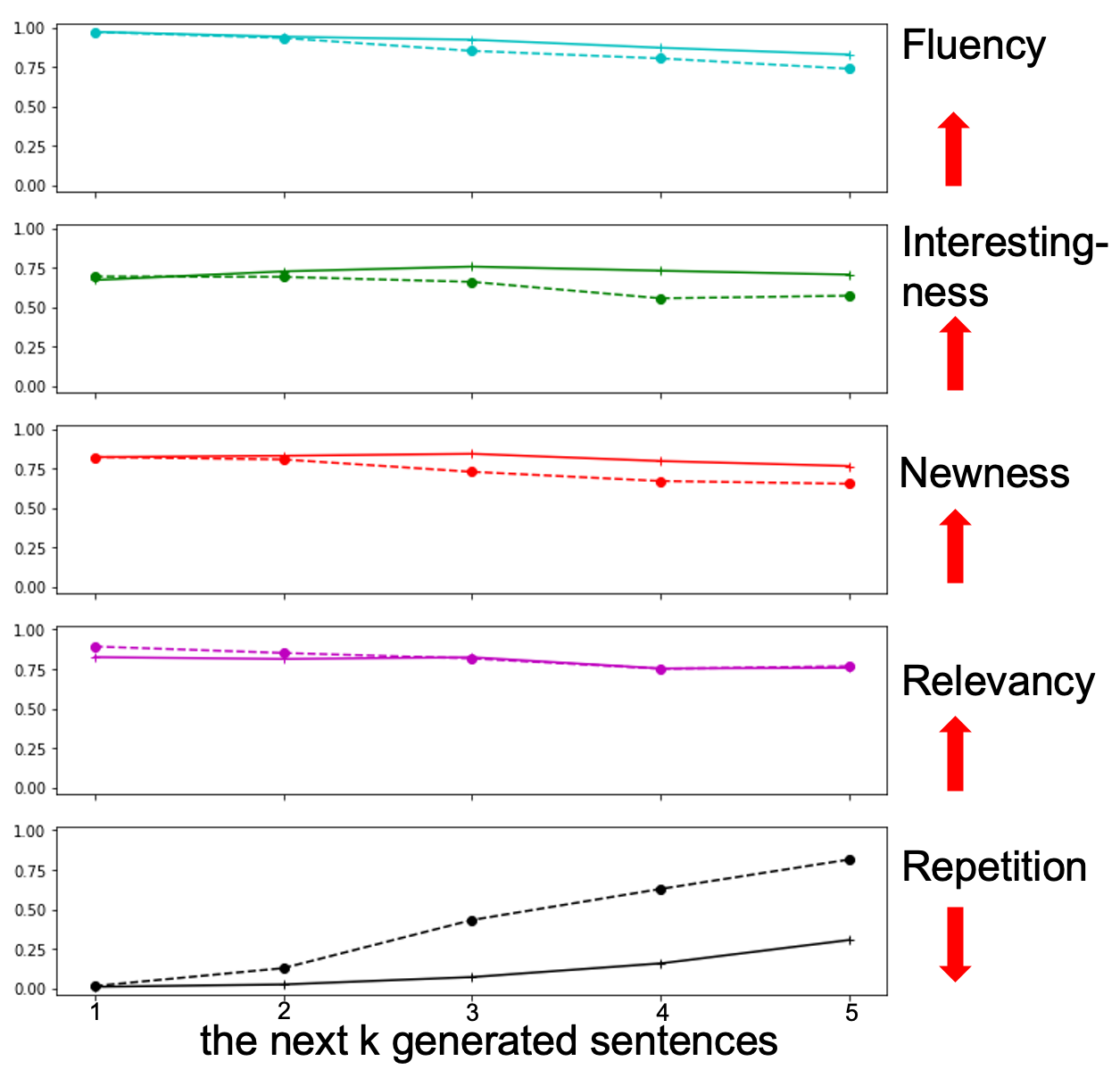}
\caption{Human evaluation results on the next 1 to 5 sentences generated without (dashed lines) and with (solid lines) attention modulation ($1000$ samples).} 
\label{fig:roc_human_result}
\end{figure}

\begin{table*}[t]\footnotesize
\centering
\resizebox{\linewidth}{!}{%
\begin{tabular}{ll|rr|rr|rrr|r}
\toprule
     &      & \textbf{R-2}             & \textbf{R-L}            & \textbf{B-3}           & \textbf{B-4}           &  \textbf{Meteor}      &   \textbf{CIDEr}    & \textbf{SPICE}      & \textbf{Coverage}       \\ \midrule
 \multirow{2}{*}{greedy} &  w/o AttnM   & 14.06        & 34.13        & 26.19        & 17.92        & 25.82   & 10.81  & 22.14  & 76.50\\ 
 &  w/ AttnM (ours)  & \textbf{14.71}        & \textbf{35.15}        & \textbf{27.53}        & \textbf{18.91}        & \textbf{26.23}   & \textbf{11.61} & \textbf{23.22}  & \textbf{79.18} \\ \midrule
\multirow{2}{*}{beam=5} &  w/o AttnM    & 16.68         & 36.92        & 32.39        & 23.36        & 26.87   & 12.24 & 22.83  & 76.99 \\
 &  w/ AttnM (ours) & \textbf{17.14}         & \textbf{38.23}        & \textbf{33.92}        & \textbf{24.03}        & \textbf{27.48}   & \textbf{12.88} & \textbf{24.24}  & \textbf{81.27}\\ \midrule

  \multirow{2}{*}{beam=10} &  w/o AttnM    & 17.25         & 37.37        & 33.81        & 24.39        & 27.51   & 12.58 & 23.24  & 78.68 \\

 &  w/ AttnM (ours)  & \textbf{17.59}         & \textbf{38.71}        & \textbf{35.72}        & \textbf{25.93}        & \textbf{27.71}   & \textbf{13.32} & \textbf{24.36}  & \textbf{81.24}\\ \midrule

\multirow{3}{*}{beam=20}& \citet{lin-etal-2020-commongen} & 16.85&39.01&33.92&23.73&26.83&12.19&23.57&79.09\\
&  w/o AttnM    & 17.98        & 38.07        & 35.14        & 25.61       & 27.63   & 12.90 & 23.28  & 79.62 \\
 &  w/ AttnM (ours)  & \textbf{18.11}         & \textbf{39.32}        & \textbf{36.69}        & \textbf{26.80}        & \textbf{28.02}   & \textbf{13.71} & \textbf{23.94}  & \textbf{81.85}\\ \bottomrule
\end{tabular}}
\caption{CommonGen test results of the fine-tuned GPT2-L  w/ or w/o \alg{} based on different decoding algorithms. }\label{tab:commongen_greedy}
\end{table*}

\begin{table}[t]
\centering
\resizebox{\columnwidth}{!}{%
\begin{tabular}{c|cccc|c}
\toprule
& \textbf{\small B-4} &  \textbf{\small R-L} & \textbf{\small Meteor}       & \textbf{\small CIDEr}  &\textbf{\small Human}   \\
 \midrule
w/o AttnM & 13.51       & \textbf{18.29}   & 13.18   & 47.69    &14\%      \\
w/ AttnM  (ours) & \textbf{13.52}& 18.01  & 13.18            & \textbf{48.20}  &\textbf{33\%}         \\ \bottomrule
\end{tabular}}
\caption{Evaluations of the fine-tune GPT2-L on $\alpha$NLG  using greedy decoding.}\label{tab:anlg_auto_result}
\end{table}

\subsection{ROCstories}\label{subsec:result_roc}
 Table \ref{tab:roc_result} indicates that \alg{} significantly reduces repetition in narrative generation, while increasing the relevancy of generated sentences to the original story.  We can observe a vast improvement in the number of unique generated tokens using \alg{}, indicating a reduced repetition rate (confirmed by the \% number of repeated sentences in the next five sentence generated -- 35.43 vs. 17.49 for our approach). 
This intuition is confirmed by our human evaluation in Figure~\ref{fig:roc_human_result}, where the GPT2-L with \alg{} produces sentences that are more fluent, more interesting, more novel, and less repetitive than the original decoder. Furthermore, we note that the difference in performance across these evaluation categories generally increases as the number of generated sentences increases, indicating less sensitivity to long-form degeneration.

\subsection{$\alpha$NLG}


Table \ref{tab:anlg_auto_result} presents the automatic and human evaluation results on $\alpha$NLG. We can see that our model performs similarly with and without attention modulation in terms of automatic evaluation. However, our human evaluation results in the last column show that overall, the human judges prefer the explanations produced using attention modulation significantly more than those of the original model. With 100 samples generated, 33$\%$ of the time, human judges select explanations generated with \alg{} as more plausible. In contrast, explanations from the original model are only preferred 14$\%$ of the time.

\begin{table*}[h]
\centering
\resizebox{\linewidth}{!}{%
\begin{tabular}{ll|rr|rr|rrr|l}
\toprule
 \textbf{Training size}    &   \textbf{Method}   & \textbf{R-2}             & \textbf{R-L}            & \textbf{B-3}           & \textbf{B-4}           &  \textbf{Meteor}      &   \textbf{CIDEr}    & \textbf{SPICE}      & \textbf{Coverage}       \\ \midrule
  \multirow{2}{*}{10} &  w/o AttnM   & 2.15       & 16.07       & 6.39        & 3.63       & 10.02   & 1.23 & 5.07  & 27.12\\ 
 &  w/ AttnM  (ours) & \textbf{3.38}        & \textbf{18.89}        & \textbf{7.24}        & \textbf{3.48}        & \textbf{12.47}   & \textbf{1.63} & \textbf{7.21}  & \textbf{36.55} \quad \color{red}{$\uparrow$ 9.43} \\ \midrule
 
\multirow{2}{*}{1000} &  w/o AttnM   & 7.61       & 27.03       & 14.01       & 7.38       & 20.67   & 6.40 & 16.79  & 62.33\\ 
&  w/ AttnM  (ours) & \textbf{8.54}        & \textbf{27.97}        & \textbf{15.51}        & \textbf{8.78}        & \textbf{22.13}   & \textbf{6.68} & \textbf{18.02}  & \textbf{69.31} \quad \color{red}{$\uparrow$ 6.98} \\ \midrule

 \multirow{2}{*}{10000} &  w/o AttnM   & 10.70       & 30.06        & 17.40        & 10.02        & 23.40   & 7.15 & 20.20  & 73.39\\ 
 &  w/ AttnM (ours)  & \textbf{11.53}        & \textbf{30.70}        & \textbf{18.43}        & \textbf{11.12}        & \textbf{24.42}   & \textbf{7.29} & \textbf{21.33}  & \textbf{78.74} \quad \color{red}{$\uparrow$ 5.35} \\ \midrule

 \multirow{2}{*}{full ($\sim$39k)} &  w/o AttnM   & 14.06        & 34.13        & 26.19        & 17.92        & 25.82   & 10.81  & 22.14  & 76.50\\ 
 &  w/ AttnM (ours)  & \textbf{14.71}        & \textbf{35.15}        & \textbf{27.53}        & \textbf{18.91}        & \textbf{26.23}   & \textbf{11.61} & \textbf{23.22}  & \textbf{79.18} \quad \color{red}{$\uparrow$ 2.68} \\  \bottomrule
\end{tabular}} \caption{CommonGen test results of the fine-tuned GPT2-L w/ or w/o \alg{} trained on different size of training examples (greedy decoding).}\label{tab:commongen_few_shot}
\end{table*}

\subsection{CommonGen}
\begin{table}[ht]
\centering
\begin{tabulary}{\textwidth}{C|CCC}
\toprule
     & \textbf{Fluency} & \textbf{Quality} & \textbf{Overall} \\ \midrule
w/o AttnM  & 85.07   & 39.30   & 44.77   \\
 w/ AttnM (ours) & \textbf{89.55}   & \textbf{48.76}   & \textbf{52.73}   \\ \bottomrule
\end{tabulary} 
\caption{Human evaluations  of the fine-tuned GP2-L w/o or w/ \alg{} on CommonGen.} 
\label{tab:commongen_human_eval}
\end{table}
Table \ref{tab:commongen_greedy} shows the automatic evaluation results on the CommonGen dataset. We separate different settings of decoding algorithms in blocks. By injecting the prior -- the model should put more attention on uncovered concepts -- into the GPT2-L with \alg{},  we can improve the text generated in every automatic measure significantly. Interestingly, despite our attention-reweighted decoder only encouraging coverage, we see all the other measures such as ROUGE, BLEU, METEOR, CIDEr, SPICE improve, as well. 

These improvements also hold when we use a different base decoding algorithm, such as beam search. Again, the performance improvement for using \alg{} is significant over all measures. Thus, unlike decoding algorithms that improve downstream tasks through truncation of the sampling distribution, we directly re-calibrate the token-level probabilities predicted by the model by altering attention patterns in the Transformer blocks during inference.

We also conduct a human evaluation to check whether this improvement in the automatic metrics transfers to human judgments. In Table \ref{tab:commongen_human_eval}, we see that our \alg{} algorithm significantly outperforms the original inference model on every measure -- from fluency, quality, and overall performance.

\subsection{Vast improvements on few-shot learning}
Table~\ref{tab:commongen_few_shot} presents the results of \alg{} on GPT2-L  that are fine-tuned on different numbers of training examples from CommonGen. We observe the improvements on all measures are more prominent when the fine-tuning data size is small. For example, adding \alg{} can improve coverage by 9.43\%  on the GPT2-L fine-tuned with only 10 examples.
This not only validates that priors we injected into the model are suitable for improving the downstream task performance, but also shed lights to use \alg{} on different few-shot learning scenarios where the number of training examples is limited. 

\section{Related Work}
We propose to use attention modulation to heuristically  re-balance sentence-level attention for neural text degeneration.  At least three domains of work are closely related to our proposal, namely, attention pattern analysis, work that focuses on changing or approximating learned attention patterns, and work for countering neural text degeneration. 

\paragraph{Attention analysis:} 
Previous work has investigated the attention patterns within the local context of a sentence. These works highlighted that attention patterns in Transformers implicitly encode syntactic information such as dependency relations \citep{htut2019attention}, and part-of-speech tags \citep{vig-belinkov-2019-analyzing,raganato-tiedemann-2018-analysis}. Other works observed that attention patterns can provide explanations \citep{wiegreffe2019attention} or coarse word alignments in machine translation \citep{zenkel2019adding,kobayashi-etal-2020-attention}. 
In contrast to these works, we analyze sentence-level attention patterns for neural text degeneration, and propose to directly modify the attention computation to reduce it.  

\paragraph{Alternative attention:}
Many works have been proposed to change attention mechanisms to optimize their O($n^2$) complexity. Some promising directions in this space include sparse attention mechanisms \citep{beltagy2020longformer,zaheer2020big} and linearized attention \citep{choromanski2020rethinking}. These alternative attention mechanisms require training the model and are used as replacements to the original attention mechanism for fast training or reduced computation.  
Our work is fundamentally different as we seek to inject priors into the standard attention mechanism during inference (without re-training the model). 

\paragraph{Neural text degeneration:}
Previous works seek to solve neural text degeneration by changing the training objective to reduce the likelihood of common tokens \citep{welleck2020neural}, or modifying the decoding algorithm by truncating the sampling distribution \citep{holtzman-etal-2018-learning,holtzman2020curious}.  
Specifically, \citet{welleck2020neural} introduce an additional training loss that reduces the likelihood of common tokens. 
\citet{holtzman-etal-2018-learning,holtzman2020curious} propose stochastic decoding algorithms with truncation of the sampling distribution. 
Our work is orthogonal to these methods by injecting priors into the model's attention computation during inference.

\section{Conclusions and future work}
Neural language models often exhibit degeneration: the output texts are repeated, bland, and inconsistent. Our empirical analyses show that neural text degeneration may be associated with insufficient learning of task-specific characteristics by the attention mechanism. We propose a simple but effective module -- \alg{} -- that can inject priors for better generation through re-balancing the attention weights during inference. Results on three different narrative and commonsense generation tasks indicate that \alg{} can reduce repetition and enhance commonsense reasoning while maintaining fluency and coherence.


\section*{Acknowledgements}
This research was supported in part by NSF (IIS-1714566), DARPA MCS program through NIWC Pacific (N66001-19-2-4031), the Canada CIFAR AI Chair program, the Natural Sciences and Engineering Research Council of Canada (NSERC), Intel Labs Cognitive Computing Research, and the Allen Institute for AI. 
Computations on \url{beaker.org} were supported in part by credits from Google Cloud. 

\bibliographystyle{acl_natbib}
\bibliography{anthology,acl2021}
\clearpage
\appendix

\appendix \label{sec:appendix}
\section{Appendices}

\begin{figure}[t]
\centering
\includegraphics[scale=0.53]{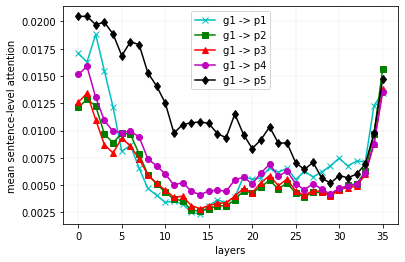}
\caption{Sentence-level attention distribution 
across different layers in GPT2-L. The result is aggregated by computing the mean sentence-level attention from the next generated sentence to the five sentences in the prompt of ROCStories development set. Lower number represents lower layers in the Transformer.}
\label{fig:sentence_level_attention_portion}
\end{figure} 
\begin{figure}[t]
\centering
\includegraphics[scale=0.55]{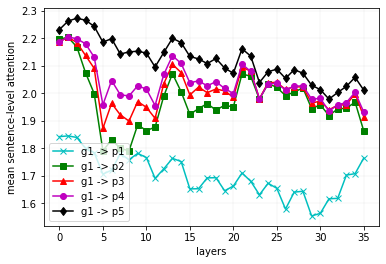}
\caption{Attention entropy of each sentence in the prompt aggregated over the ROCStories dev. set. In the
first sentences, there are particularly high-entropy attention heads that
produce bag-of-vector-like representations.}
\label{fig:mean_entropy}
\end{figure} 

\subsection{How does a language model use attention to model a multi-sentence prompt?} \label{appendix:multi_sentence_prompt}

\paragraph{Sentence-level attention portion}
To reveal which part of context -- near or distant history -- are important for context representation,  we compute aggregated mean sentence-level attention (Eqn.5 in the main text) each prompt sentence $p_i$ received, while generating the sentence $g_1$ after the prompt. We observe from Figure \ref{fig:sentence_level_attention_portion} that GPT2-L mostly attends to the nearest sentence ($p_5$) during the generation. This effect is especially prominent in the early and middle layers. In the late layers, the attention from different sentences evens out. This observation is consistent with previous analysis of attention patterns within sentences such that deeper layers focus on longer-range context \citep{vig-belinkov-2019-analyzing}.

\paragraph{Sentence-level attention entropy}
\Citet{khandelwal-etal-2018-sharp} observed that LSTM represent distant context as topics; only a few token in the distant context are used to compute the context representation. We check whether this observation also holds on Transformer-based models by computing attention entropy.  This sentence-level attention entropy of $p_i$ based on the attention from $g_1$ to $p_i$ at layer $l_m$ over a corpus $X$ is defined as: 

\begin{equation}
    \label{eq:sent_sent_attn_entropy}
    E_{A}(g_1,p_i,l_m)=-\frac{
    \sum\limits_{x \in X} 
    \sum\limits_{h \in l_m} 
    \sum\limits_{j=1}^{|p_i|}
    \sum\limits_{k=1}^{|g_1|}
    \alpha^{h}_{j,k}\text{log}(\alpha^{h}_{j,k})
    }
    {
  |X|\cdot |H| \cdot|p_i|\cdot |g_1|
    }
\end{equation}
where $h$ is a head in layer $l_m$ and $\alpha^{h}_{j,k}$ is the attention weight from $x_j \in p_i$ to $x_k \in g_1$ for $h$. Figure \ref{fig:mean_entropy} shows a clear separation of entropy over different sentences in the prompt, where more distant sentences have lower entropy values. 
This suggests that LMs  only modelling  distant sentences as topics -- attention over key words being a proxy. 
\begin{table}[t]
\centering
\begin{tabular}{ p{0.3\linewidth}|ccc}
\toprule
          & \textbf{train}             & \textbf{dev.}  & \textbf{test} \\ \midrule
ROCStories &      39498   &  5269   &   7899   \\ 
$\alpha$NLG   & 169,654   & 1,532 & 3,059 \\ 
CommonGen  & 67,389  & 4,018 & 7,644 \\ \bottomrule
\end{tabular}\caption{Dataset Statistics}\label{tab:datasets}
\end{table}

\subsection{Hyperparameters}
 \begin{table*}[!ht]
    \centering
    \begin{tabular}{p{0.2\linewidth} |p{0.1\linewidth}| p{0.6\linewidth}}
    \toprule
      propmt & attention reweight & Generated sentence with attention-decoding \\\midrule
field. stand. look. = & (\textcolor[HTML]{00a65a}{1},\textcolor[HTML]{e74c3c}{3},\textcolor[HTML]{f39c12}{2}) & A man \textcolor[HTML]{e74c3c}{stands} \textcolor[HTML]{f39c12}{looking} at a sign in a \textcolor[HTML]{00a65a}{field}.\\ 
field. stand. look. = & (\textcolor[HTML]{00a65a}{1},\textcolor[HTML]{f39c12}{2},\textcolor[HTML]{e74c3c}{3})  & He \textcolor[HTML]{e74c3c}{looks} up and sees a group of people \textcolor[HTML]{f39c12}{standing} in the \textcolor[HTML]{00a65a}{field}.\\ 
field. stand. look. = & (\textcolor[HTML]{f39c12}{2},\textcolor[HTML]{e74c3c}{3},\textcolor[HTML]{00a65a}{1}) & He \textcolor[HTML]{e74c3c}{stands} in the middle of the \textcolor[HTML]{f39c12}{field}, \textcolor[HTML]{00a65a}{looking} down at the \textcolor[HTML]{e74c3c}{stands}.\\ \bottomrule
    \end{tabular}
    \caption{ An example of \alg{} with different attention reweighting functions on CommonGen dev set. Only by redistributing the sentence-level attention during inference, we can generate sentences following the desired order specified in the attention reweighting function.}
    \label{tab:example_different_reweight}
\end{table*}

\begin{table*}[t]\footnotesize
\centering
\resizebox{\linewidth}{!}{%
\begin{tabular}{ll|rr|rr|rrr|r}
\toprule
     &      & \textbf{R-2}             & \textbf{R-L}            & \textbf{B-3}           & \textbf{B-4}           &  \textbf{Meteor}      &   \textbf{CIDEr}    & \textbf{SPICE}      & \textbf{Coverage}       \\ \midrule
\multirow{3}{*}{beam=20}
& w/o AttnM    & 17.98        & 38.07        & 35.14        & 25.61       & 27.63   & 12.90 & 23.28  & 79.62 \\
 & w/ AttnM   & 18.11        & 39.32        & 36.69       & 26.80        & 28.02   & 13.71 & 23.94  & 81.85\\ 
 & w/ AttnM$_{pm}$   & \textbf{19.78}         & \textbf{41.17}        & \textbf{38.28}        & \textbf{28.36}        & \textbf{30.97}   & \textbf{15.38} & \textbf{28.24}  & \textbf{91.89}\\ \bottomrule
\end{tabular}}
\caption{Results on CommonGen test set with beam search and permutations defined in \ref{appendix:permutation}.}\label{tab:commongen_permutation}
\end{table*}

 Attention modulation can be applied to any layer and any head in the Transformer based on our implementation. However, the weights learned by different heads in a particular layer have a large variance \citep{vig-belinkov-2019-analyzing} and are subject to change from different training sessions. Therefore, we only reweight attention on all heads in different layers, where what layers are re-weighted are hyperparameters. We choose to reweight the consecutive layers from a starting layer $l_{s}$ to  an end layer $l_{e}$ and performed a grid search on different layer ranges. For the start layer, we experimented with $l_s \in \{0,4,8,12,16,20,24,28,32\}$; for the end layer, we experimented with $l_e \in \{4,8,12,16,20,24,28,32,36\}$. The reweighting layers are chosen based on the validation set performance. On ROCstories, the GPT2-L are re-weighted with $l_s =8$ and  $l_e =32$; On $\alpha$NLG,  the GPT2-L are re-weighted with $l_s =12$ and  $l_e =32$; On CommonGen, the GPT2-L are re-weighted with $l_s =24$ and  $l_e =32$.
 
 \subsection{Attention modulation and generation order}
 \label{appendix:permutation}
 CommonGen provides the concept set in a random order, where models need to perform a relational commonsense reasoning to find the optimal order of them for generating a plausible sentence. We found that \alg{}  provide signals for generation order given different reweighting functions (examples in Table \ref{tab:example_different_reweight}). In this experiment, we guide the generation order by providing different initialization weights in the reweighting functions. We enforce different \alg{} weights based on the order we want the concepts to be generated.  For examples, row 1 in table \ref{tab:example_different_reweight} means the concepts of $\textsc{(field, stand, look)}$ are initialized to be re-weighted by scales of (1,3,2).

This interesting finding motivates us to conduct a permutation experiment on CommonGen. For a $k$ concept set, we initialize the \alg{} weights based on the permutations of 1 to $k$ ($k$! permutations in total) and generate $k$! sentences with \alg{}. We then select the generation that covers the most concepts\footnote{If there is a tie, we choose the shorter generation.} from these $k$! generations as output. We call this method "\alg{} with permutation". Table \ref{tab:commongen_permutation} presents the results of \alg{} with permutation. 
We see that just by enforcing the order in \alg{}, the coverage can be improved by another 10\%. 

 \subsection{Human evaluation details}
\begin{figure*}[ht]
\centering
\includegraphics[scale=0.6,angle =270 ]{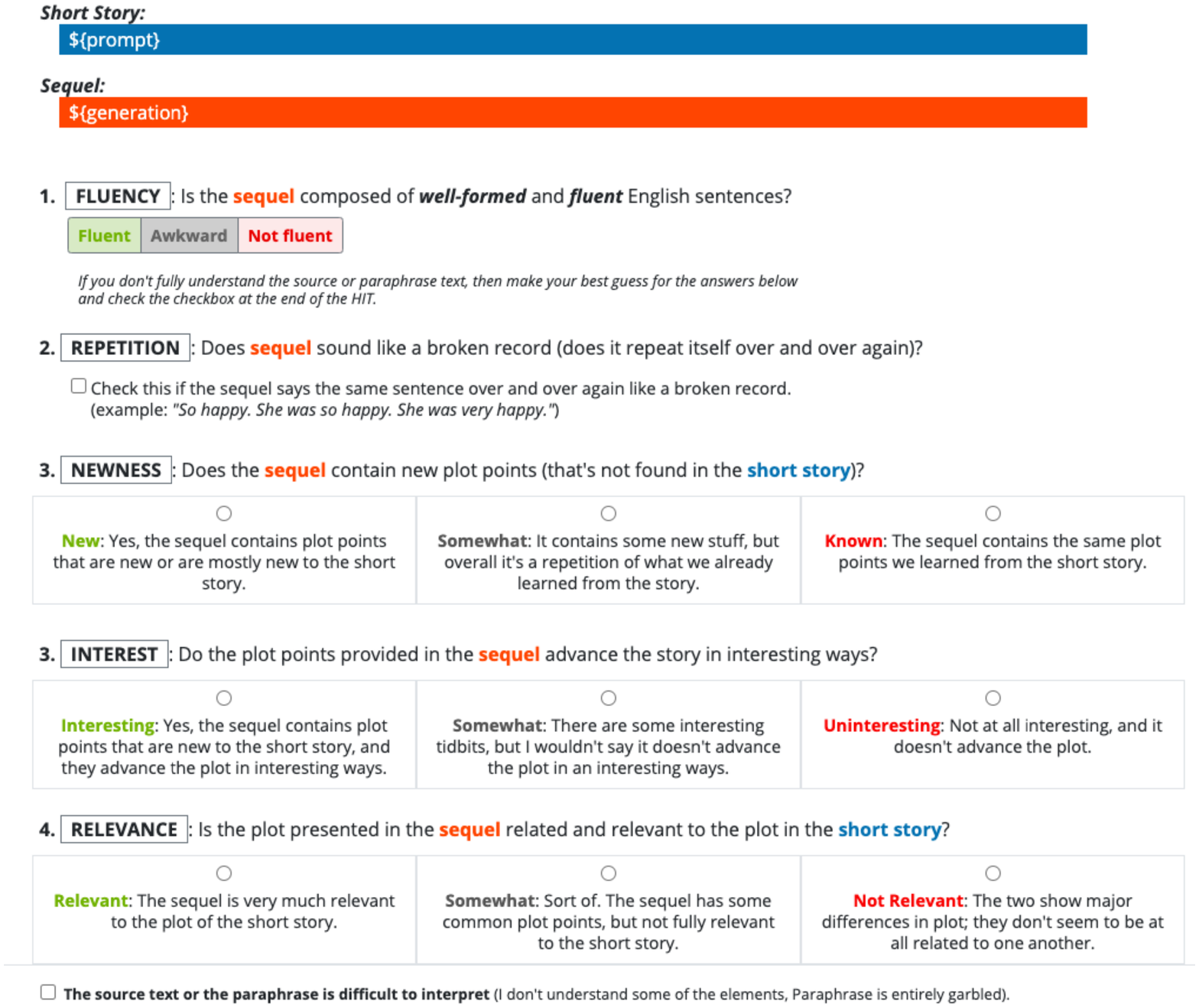}
\caption{\small Mechanical Turk template used to evaluate ROCstories generations.}
\label{fig:roc_template}
\end{figure*}

\begin{figure*}[t]
\centering
\includegraphics[scale=0.525]{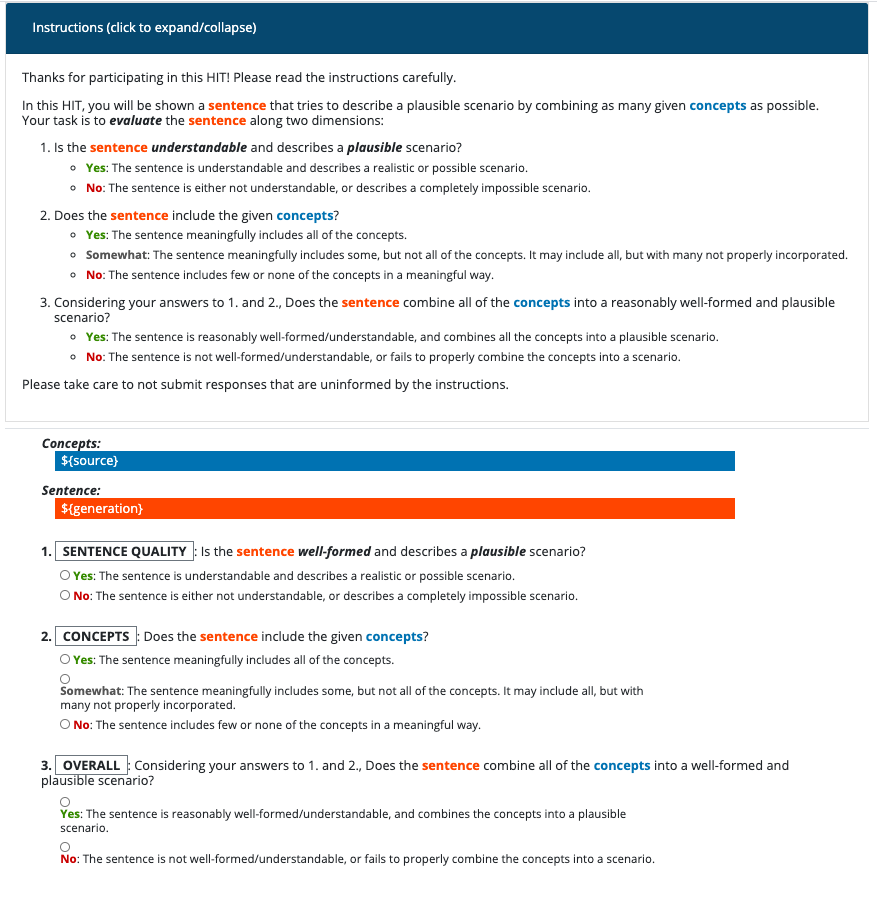}
\caption{\small Mechanical Turk template used to evaluate CommonGen generations.}
\label{fig:commongen_template}
\end{figure*}

 Figure \ref{fig:roc_template} and \ref{fig:commongen_template} show the evaluation templates  for tuckers for Rocstories and CommonGen\footnote{omitted  $\alpha$NLG template due to space limitation.}, respectively. On ROCstories, the inter-annotator agreements is 0.743 and fleiss kappa is 0.229 averaged over the five dimensions on $200 \times 5$ samples. The inter-annotator agreements are 84.6\%, 72.2\%, 76.1\%, 78.5\%, and 59.9\% for fluency, interestingness, newness, relevance, and repetition. On $\alpha$NLG, the human evaluation results are 33\%, 14\%, 10\%, 32\%, 11\% for preferring outputs with \alg{}, without \alg{}, both, neither and no agreement. The inter-annotator agreements is 0.835 and kappa is 0.439 over 100 samples.
 On CommonGen, the inter-annotator agreements  are 77.94\%, 71.64\%, and 67.50\% for fluency, quality,  and overall on 200 samples.


\begin{table*}[ht]
    \centering
     \resizebox{\linewidth}{!}{%
    \begin{tabular}{p{0.1\linewidth}| p{0.9\linewidth}}
      prompt & Jen was always afraid of flying. She needed to fly to Japan for an important meeting. Jen took some Dramamine and boarded her plane. Jen dozed off shortly after the plane took off. \\\hline
w/o AttnM & Jen woke up to a huge headache. She had to cancel her trip to Japan. Jen was glad she had taken Dramamine. She was able to fly safely. Jen was glad she had taken Dramamine.\\ \hline
w/ AttnM & She woke up to a loud bang and felt sick. She had to sit in the back of the plane. She was so scared she couldn't even get out of her seat. Jen was so embarrassed she never flew again. She was so scared she never flew again.\\ \hline
    \end{tabular}}
    \vspace{-1mm}
    \caption{ A generation example (greedy decoding) without and with \alg{} on ROCStories. Adding \alg{} during inference \textbf{improves the commonsense reasoning and consistency of the generation}.}
    \label{tab:example_roc}
\end{table*}

\begin{table*}[ht]
    \centering
    \resizebox{\linewidth}{!}{%
    \begin{tabular}{p{0.1\linewidth} | p{0.9\linewidth}}
      prompt & Neil had been journeying through Asia. But he contracted malaria on a Thai island, and had to be flown home for treatment. \\\hline 
 
w/o AttnM &  He was so sick that he couldn't go to the hospital. He was so sick that he couldn't go to the hospital. He was so sick that he couldn't go to the hospital. He was so sick that he couldn't go to the hospital. He was so sick that he couldn't go to the hospital.
 \\ \hline
w/ AttnM & 
He was very sick and had to be hospitalized. He was in the hospital for a week.
He recovered and was released. Neil was very happy to be home. He was able to get better and was able to go back to his home country. \\ \hline
\end{tabular}}
\vspace{-1mm}
\caption{Example (greedy decoding) of the fine-tuned GPT2-L without and with \alg{} on story completion. Adding \alg{} during inference significantly \textbf{reduces the sentence-level repetition}.}
\label{tab:story_repetition}
\end{table*}

\begin{table*}[ht]
    \centering
    \begin{tabular}{p{0.12\linewidth} | p{0.25\linewidth} | p{0.55\linewidth}}
    \toprule
      &propmt & Generated sentence with attention-decoding \\\midrule
w/o AttnM &run. team. field. drill. = &  person \textcolor[HTML]{e74c3c}{runs} a \textcolor[HTML]{e74c3c}{drill} during a practice at training camp. \\ 

w/ AttnM. &run. team. field. drill. = &
person \textcolor[HTML]{e74c3c}{runs} a \textcolor[HTML]{e74c3c}{drill} during a training session with his \textcolor[HTML]{e74c3c}{team}.\\ \midrule
w/o AttnM &use. tool. piece. metal. = & \textcolor[HTML]{e74c3c}{tool} or \textcolor[HTML]{e74c3c}{piece} of \textcolor[HTML]{e74c3c}{metal} \textcolor[HTML]{e74c3c}{used} in manufacturing.\\
w/ AttnM &use. tool. piece. metal. =& \textcolor[HTML]{e74c3c}{piece} of \textcolor[HTML]{e74c3c}{metal} \textcolor[HTML]{e74c3c}{used} to make \textcolor[HTML]{e74c3c}{tools}.\\ \bottomrule

    \end{tabular}
    \caption{ Examples produced by GPT2-L without  and with  \alg{}.   Use \alg{} would have higher concept coverage (details in Table 3 in the main text with 5\% coverage improvements on all decoding algorithms we tested);}
    \label{tab:example_concept_coverage}
\end{table*}
\vspace{-2cm}


\end{document}